\documentclass{article}[11pt]

\usepackage{bbm}
\usepackage{bm}

\newcommand{\vd}{\bm{d}}

\newcommand{\vh}{\bm{h}}

\newcommand{\vu}{\bm{u}}

\newcommand{\vx}{\bm{x}}
\newcommand{\vy}{\bm{y}}

\newcommand{\vU}{\bm{U}}

\newcommand{\vtheta}{\bm{\theta}}

\newcommand{\ignore}[1]{}

\newcommand{\vsp}{\vspace{0.5ex}}
\newcommand{\ds}{\displaystyle}
\newcommand{\bu}{\bm{u}}

\newcommand{\norm}[1]{\left\lVert#1\right\rVert}
\usepackage{amsmath,amsthm,amssymb,graphicx}
\usepackage[lined,boxed,commentsnumbered]{algorithm2e}
\usepackage{pgfplots}
\usepackage{bm}
\usepackage{hyperref}
\usepackage{natbib}
\usepackage{comment}
\usepackage{booktabs}
\usepackage{subcaption}
\usepackage{tabularx}

\renewcommand{\vec}[1]{\mathbf{#1}}

\newcommand{\half}{{\textstyle\frac{1}{2}}}

\usepackage{amsmath,amssymb,amsthm}

\title{Fairness through Social Welfare Optimization}

\author{Violet (Xinying) Chen \thanks{Stevens Institute of Technology, \texttt{vchen3@stevens.edu}}, J. N. Hooker \thanks{Carnegie Mellon University, \texttt{jh38@andrew.cmu.edu}}}
\date{July 2022}

\begin{document}
\maketitle

\begin{abstract}
We propose social welfare optimization as a general paradigm for formalizing fairness in AI systems. We argue that optimization models allow formulation of a wide range of fairness criteria as social welfare functions, while enabling AI to take advantage of highly advanced solution technology. Rather than attempting to reduce bias between selected groups, one can achieve equity across all groups by incorporating fairness into the social welfare function.  This also allows a fuller accounting of the welfare of the individuals involved.  We show how to integrate social welfare optimization with both rule-based AI and machine learning, using either an  in-processing or a post-processing approach. 
We present empirical results from a case study as a preliminary examination of the validity and potential of these integration strategies. 
\end{abstract}

\section{Introduction}
Artificial intelligence is increasingly used not only to solve problems, but to recommend action decisions that range from awarding mortgage loans to granting parole.  The prospect of making decisions immediately raises the question of ethics and fairness.  If ethical norms are to be incorporated into artificial decision making, these norms must somehow be automated or formalized.  The leading approaches to this challenge include 
\begin{itemize}
    \item {\em value alignment}, which strives to train or modify AI systems to reflect human ethical values automatically, e.g. \cite{AllSmiWal05,Rus19,Gab20};
    \item {\em logical formulations} of ethical and fairness principles that attempt to represent them precisely enough to govern a rule-based AI system, e.g. \cite{BriArkBel06,LinMatNeb20,HooKim2018}; and
    \item {\em statistical fairness metrics} that aim to ensure that benefits are allocated equitably in the decision process, e.g. \cite{DwoHarPitReiZem12,MehMorSaxLerGal19,ChoRot20}.
\end{itemize}
Each of these approaches can be useful in a suitable context. We wish to propose, however, an alternate framework for formalizing ethics and fairness that has received less attention: 
\begin{itemize}
    \item {\em social welfare optimization}, which allows one to achieve equity or fairness by maximizing a {\em social welfare function}. 
\end{itemize}
Welfare economics has long used social welfare functions (SWFs) as a tool to measure the desirability of a given distribution of benefits and harms. A SWF is a function of the utility levels allocated to affected parties, where utility reflects a party's gain or loss as a consequence of the decisions of interest. Using a SWF motivates explicit consideration of the downstream outcomes of fairness and equity criteria. In contrast to leading notions of AI fairness that focus on eliminating disparity between groups, SWFs allow a broader perspective that emphasizes fairness in the welfare impacts of decisions. 

AI research is beginning to recognize the importance of a welfare perspective on fairness (e.g., \cite{corbett2018measure,hu2020fair}), due in part to its potential for aligning fairness concepts with social well-being. Despite this rising attention, there is no general framework for incorporating welfare-based fairness into AI systems. In this paper, we utilize social welfare optimization as the core component of one possible framework. This framework allows one to take advantage of the flexibility of SWFs to represent a wide range of fairness and equity concepts, as well as to harness powerful optimization solvers.  Optimization methods are of course already employed in AI to train neural networks, calibrate machine learning models, and the like.  Our proposal is to formalize welfare perspectives on AI fairness through social welfare optimization.

We begin below by stating some specific advantages of social welfare optimization as a paradigm for implementing equity and fairness in AI. We then introduce a bit of notation, state the basic optimization problem, and present an example of mortgage loan processing. Drawing motivation from this example, we describe a general framework for welfare-based fairness through various integration schemes between AI and optimization. We continue with a review of related work in both the operations research and AI communities, and we examine several SWFs to illustrate how they can capture a variety of fairness concepts. We then show how the mortgage example can be implemented in a simple setting, and observe that social welfare maximization can improve group parity.  We conclude by outlining possible research directions.

\subsection{Advantages of Social Welfare Optimization}
The optimization of social welfare functions offers several advantages as a framework for incorporating fairness into AI.

\begin{itemize}
\item Social welfare functions provide a {\em broader perspective on fairness} than can be achieved by focusing exclusively on bias and concepts of parity across groups.  They can represent a wide range of fairness concepts and take into account the {\em level of benefit or harm} to those affected.  There is no need to decide which specific groups should be protected, because maximizing social welfare tends to promote equity {\em across all groups}.  

\item Social welfare functions allow one to {\em balance equity and efficiency} in a principled way.  Where equity is an issue, there is often  a desire for efficiency as well.  A social welfare approach obliges one to consider how equity and utilitarian goals should be represented and balanced when one chooses the function to be maximized.  

\item Optimization models allow one to harness {\em powerful optimization methods}, which have been developed and refined over a period of 80 years or more.  A wide variety of social welfare functions can be formulated for solution by highly advanced linear, nonlinear, and mixed integer programming solvers.  Formulations are provided in \cite{CheHoo22} for the SWFs mentioned herein.

\item Optimization models offer enormous flexibility to {\em include constraints on the problem}. Decisions are normally made in the context of resource constraints or other limitations on possible options.  These can be represented as constraints in the optimization problem, as nearly all state-of-the-art optimization methods are designed for constrained optimization.  

\end{itemize}

\section{The Basic Optimization Problem} \label{sec:optprob}
The general problem of maximizing social welfare can be stated
\begin{equation}
    \max_{\vd} \big\{ W\big(\vU(\vd)\big)\;\big|\; \bm{d}\in S_{\bm{d}} \big\}
    \label{eq:SWF1}
\end{equation}
where $\bm{d}=(d_1,\ldots,d_n)$ is a vector of resources distributed across stakeholders $1,\ldots, n$, and $S_{\bm{d}}$ is the set of feasible values of $\bm{d}$ permitted by resource limits and other constraints.  $\bm{U}=(U_1,\ldots,U_n)$ is a vector of {\em utility functions}, where $U_i(\vd)$ defines the utility experienced by stakeholder $i$ as a result of the resource distribution $\vd$.  We can normally write $U_i(\vd)$ as $U_i(d_i)$, since a stakeholder's utility typically depends only on the resources allotted to that stakeholder.  Finally, $W(\vu)$ is a {\em social welfare function} that measures the desirability of a vector $\vu$ of utilities.  Problem \eqref{eq:SWF1} maximizes social welfare over all feasible resource allocations.  


\section{Example: Mortgage Loan Processing}  \label{sec:ex-loan}
We use mortgage loan processing as an example, as it is a much-discussed application of AI-based decision making.  Issues of fairness arise when an AI system is more likely to deny loans to members of certain groups, perhaps reflecting minority status or gender.  A frequently used remedy is to apply statistical bias metrics to detect the problem and adjust the decision algorithms in an attempt to solve it.  

Yet bias is only one element of a broader decision-making context.  For one thing, there is a clear utilitarian imperative.  The reason for automating mortgage decisions in the first place is to predict more accurately who will default, because defaults are costly for the bank and devastating to home buyers.  The desire for accurate prediction is, at root, a desire to maximize utility.  Furthermore, bias is regarded as unfair in large part because it reduces the welfare of a segment of society that is already disadvantaged.  An aversion to bias is, to a great degree, grounded in a desire for distributive justice in general.  All this suggests that loan decisions should be designed to achieve what we really want: efficiency and distributive justice, rather than focusing exclusively on predictive accuracy and group parity.  

The social welfare function $W$ in \eqref{eq:SWF1} should be selected to balance efficiency and equity in a suitable fashion; we consider some candidate SWFs in Section~\ref{sec:swfdef}.  The stakeholders $1,\ldots, n$ might include the loan applicants, the bank, the bank's stockholders, and the community at large. For simplicity, we focus on the loan applicants as stakeholders. The utility function $\vU$ converts a given set of loan decisions $\vd = (d_1,\ldots,d_n)$ to a vector of expected utilities $\vu=(u_1,\ldots,u_n)=\vU(\vd)$ that the stakeholders experience as a result. More specifically, we let $d_i$ denote the approved loan amount, then $0 \leq d_i \leq r_i$ where $r_i$ is the requested amount by $i$, and $d_i = 0$ means the loan request is denied.
The utility measure $u_i=U_i(d_i)$ for applicant $i$ could depend on the applicant's financial situation as well as the amount of the loan, as for example when the overall benefits for receiving a loan is greater for an applicant who is less well-off. The SWF can reflect a preference for granting loans to disadvantaged applicants even when they have a somewhat higher probability of default, so as to ensure a more just distribution of utility.  This could have the effect of avoiding bias against minority groups, but as part of a more comprehensive assessment of social welfare.

\section{A General Framework} \label{sec:framework}
Drawing motivation from the mortgage example, we formalize a general framework for designing AI systems embedded with welfare-based fairness considerations. 

\subsection*{Step 1: Specify the decision problem}
We begin by specifying the needed components of the decision problem. This step is critical for the success of later steps as it ensures we have a precise understanding of the problem scope and context. We highlight some key components that commonly exist in problem instances. Additional elements may be needed for specific problems. 
\begin{itemize}
    \item {\it Task.} The task is defined by the resources to be allocated and the action to be taken.   In the mortgage example, the bank's task is to decide upon the loan amount to be granted to each applicant. 
    \item {\it Stakeholders.} Stakeholders are individuals or groups directly or indirectly affected by the decisions.  They are the utility recipients in the social welfare model. 
    \item {\it Goals.} The goals are desirable outcomes of the decision problem and serve as guiding principles for selecting the social welfare function $W(\vu)$.  The goals can encompass total utility created as well as how it is distributed.  
    
    \item {\it Constraints.} These are restrictions that limit which actions are feasible and therefore define the domain $S_{\vd}$. A main source of restriction is the scarcity of resources.  In the mortgage example, the bank is subject to budget constraints and total risk tolerance. 
\end{itemize}

\subsection*{Step 2: Define utility and social welfare functions}

The appropriate definition of utility varies with the application.  It can sometimes be identified directly with quantity of resources allocated to an individual, in which case $U_i(d_i)=d_i$.  In other cases, it can be measured by the financial benefit provided or the negative cost incurred by resources allocated, which tends to result in linear utility functions $U_i(d_i)=\alpha_i d_i$.   In health applications, it is often measured  in quality-adjusted life years (QALYs).   When there are decreasing returns to scale, a concave nonlinear utility function $U_i(d_i)$ can be used. 

A suitable social welfare function likewise depends on the application.  In some contexts, minimizing inequality may be the overriding goal, while in others one may wish to give special priority to the disadvantaged or combine efficiency and equity in some fashion.  We provide below a sampling of SWFs that may be appropriate.  Additional SWFs are surveyed in \cite{CheHoo22}.

\subsection*{Step 3: Develop a decision model} \label{sec:step3}

Decision models for social welfare optimization can be developed for either rule-based AI systems or machine learning.

\subsubsection*{Rule-based AI}
Rule-based systems are increasingly recognized for their \mbox{capacity} to support principled and transparent AI in various application domains.  For instance, \cite{Brandom2018} observes the trend in autonomous vehicle industry whereby ``companies have shifted to rule-based AI, an older technique that lets engineers hard-code specific behaviors or logic into an otherwise self-directed system.''   \cite{KimHooDon2021} demonstrate that ethical rules can be precisely represented as logical propositions suitable for inclusion in a rule base.  

Social welfare can be incorporated in a rule-based system in either of two ways.  One uses the social welfare optimization problem to guide the selection of rules to encode directly into the AI system.  It then relies on the system to make decisions as cases arise.
In the mortgage example, the bank may pre-specify applicant classes, based on applicant financial data, and determine decision rules for these classes using a social welfare optimization model. Such a rule-based system is straightforward to use: for a new loan applicant, the bank would first identify which class the applicant belongs to, then approve the loan if the corresponding rule for the class says so and reject otherwise. 

Alternatively, one can include rules in the AI rule base that provide instructions for formulating the optimization problem and for choosing actions based on the optimal solution. This is consistent with the proposal from \cite{BriArkBel06} that one could constrain AI systems with ethical principles formalized as logic statements, such as if-then statements. In the example, the bank may consider rules that require applicants with certain features to receive reasonable prioritization, and these rules can be captured as constraints or incorporated into the objective function in the optimization model. Furthermore, when making the final loan decisions, the bank may define rules about implementing the allocation solution obtained from the optimization problem.

\subsubsection*{Machine Learning}
There are at least two ways in which social welfare optimization can be integrated with machine learning.  A \textit{post-processing} approach trains the machine to predict the information needed to formulate the welfare optimization model, whereupon the model is solved for decisions.  An \textit{in-processing} approach combines the loss function of the learning mechanism with a social welfare function, perhaps by taking a convex combination of the two.  As a result, the machine yields decisions that are already welfare sensitive, and no post-processing is necessary.  

We can make this more precise for supervised learning as follows. Suppose a training data set is $\mathcal{D} = \{(\vx_i, y_i) \}_{i = 1}^n$ where $\vx_i$ is the feature vector and $y_i$ is the true label, then a supervised learning method trains a model $h$ with the accuracy of the predicted information $\{h(\vx_i)\}$ as the primary goal. The ML literature has studied a large number of formats for $h$, ranging from a simple functional form in logistic regression and support vector machine to more complex structures like decision trees and neural networks. 

In the post-processing approach, the prediction step focuses solely on accuracy through minimization of a loss function $\mathcal{L}$.  The prediction and decision steps can be formalized as
\begin{gather*}
\begin{aligned}
    & \textit{Prediction step: } h^* = \text{argmin}_{\vh} \mathcal{L}(\vh,\mathcal{D}); \\
    & \textit{Decision step: } \vd^* = \text{argmax}_{\vd} \big\{W\big(\vU(\vd)\big): d_i = d\big(\vx_i, h^*(\vx_i)\big)\big\}.
\end{aligned}
\end{gather*}
In the mortgage example, we might suppose that $h(\vx_i)$ is the probability that applicant $i$ will repay the loan if it is granted.  Then if $d_i$ is the loan amount, we could set $U_i(\vd)=U_i(d_i)=h^*(\vx_i)d_i$.  It is notable that all supervised learning methods are suitable for this type of integration, and the decision maker has the flexibility to choose the ML method that fits the problem's context and computational requirement.   

The {\em in-processing} approach can be viewed as a type of in-processing fair ML method, but distinguished from most of the literature in that we modify the standard accuracy objective in a training model to reflect social welfare. A simple scheme is to take a weighted sum of training loss and negative social welfare. The if $h(\vx_i)$ is the predicted decision for individual $i$, we have
\begin{align*}
    & \textit{Prediction/decision step: } 
    h^* = \text{argmin}_h \big\{ \mathcal{L} (h,\mathcal{D}) 
    - \lambda W(\vU(\vd)) : d_i=h(\vx_i) \big\}
\end{align*}
The success of this type of integration is contingent on whether the resulting training model can be solved efficiently.

As a final remark, we briefly note the integration potential for two other core machine learning methods, unsupervised learning and reinforcement learning. Fairness has been studied in both methods, but progress is much more limited relative to fair supervised learning. With respect to unsupervised learning, we can easily apply post-processing integration to clustering methods (for example) and utilize the trained clusters as input to specify the optimization problem. In the loan example, the bank can use clustering algorithms to categorize financial profiles that will play a role in the optimization formulation. Recent work in fair clustering, e.g.\ \cite{abraham2019faircluster} and \cite{deepak2020representativity}, explore an in-processing strategy to extend K-means clustering to include fairness considerations by adding a fairness component to the usual K-means objective function. 

In reinforcement learning (RL), the goal is to search for a reward-maximizing policy in a dynamic environment that is typically modelled as a Markov Decision Process. Defining and achieving fairness in RL is more challenging due to the sequential and dynamic structure. \cite{weng2019RLfairness,siddique2020fairRL} propose a novel framework for fair multi-objective reinforcement learning based on welfare optimization. The key component of their proposal is to replace the standard reward objective with a particular social welfare function on the reward distribution. This precisely exemplifies in-processing integration and demonstrates the potential of social welfare optimization for seeking fairness in RL. In addition, \cite{chohlas2021learning} propose a multi-armed bandit inspired framework for learning a fair policy, where fairness is defined with respect to the consequences of the decisions. A core element of their approach is a social welfare optimization problem, where the objective characterizes a trade-off between fairness and efficiency.

\section{Previous Work} \label{sec:util-swf}
Social welfare optimization is already fairly well established in the operations research literature, and it is beginning to attract interest in the AI community. We view AI in a broad sense to refer to all models and algorithms that can support decision-making. Our proposal is that AI expand these initial efforts into a general research program for formulating fairness.  We review here some of the previous work in both literatures. 

An excellent survey of equity models used in operations research is provided by \cite{Karsu2015}. We mention a few examples that combine equity and efficiency.  Bandwidth allocation in telecommunication networks is a popular application studied in early works on fair resource allocation (\cite{luss1999equitable,ogryczak2002equitable,Ogryczak2008}). For problems in this domain, a standard setup is to interpret bandwidth as utility and define a SWF that is consistent with a Rawlsian maximin criterion.  The corresponding optimization problem seeks equitable allocations that optimize the worst performance among activities or services that compete for bandwidth. 
Project assignment is another application where fairness is often relevant, as the involved stakeholders may have different preferences over projects. For instance, \cite{chiarandini2019handling} work with a real-life decision to assign projects to university students. They use student rankings of projects as utilities and study a variety of SWFs that capture different fairness-efficiency balancing principles. 
Fair optimization has also received attention in humanitarian operations. \cite{EisTzu19} study an important logistical challenge in food bank operations, food pickup and distribution. They design a routing resource allocation model to seek both fair allocation of food to different agencies and efficient delivery of as much food as possible. The utilities of agencies are measured by the amount of food delivered.  An SWF is selected to combine utility and the Gini coefficient.  \cite{Sibel2019inequity} consider a disaster preparation task of selecting shelter locations and assigning neighborhoods to shelters. They choose a SWF that combines the Gini coefficient with neighborhood utilities based on the travel distances to their assigned shelter. 

Recent AI research has developed efficient algorithms that take fairness into account. This effort is not directly comparable to our proposal in that it develops algorithms to solve specific problems that have a fairness component, rather than formulating optimization models that can be submitted to state-of-the-art software.  Algorithmic design tasks are often associated with fair matching decisions, such as kidney exchange \cite{McElfresh2018}, paper-reviewer assignment in peer review \cite{SteShaSin2018}, or online decision procedures for a complex situation such as ridesharing \cite{nanda2020balancing}. 

Fair machine learning is a rapidly growing field in recent years. Fair ML methods in literature can be categorized as pre-, in-, or post-processing, which respectively seek fairness by modifying standard ML methods before, during, or after the training phase. The majority of fair ML methods seek to eliminate bias and discrimination in standard ML models, via fairness notions that measure certain type of disparity in the generated predictions. Many of these methods rely on optimization in the fairness-seeking components. Pre-processing methods can use optimization models to find the best data modifications to the training data to prevent bias and disparity (see e.g. \cite{zemel2013learning, calmon2017optimized}). Similarly, post-processing methods can use optimization models to determine the optimal tuning rules to adjust the predictions generated from the trained model to seek fairness (see e.g. \cite{hardt2016equality, alabdulmohsin2020fair}). Moreover, fairness through optimization fits naturally into in-processing methods, which modify standard ML models by adding fairness constraints or including fairness components in objective function (see e.g. \cite{zafar2019fairness,olfat2018spectral,donini2018empirical}.

In contrast to this dominant statistical view of fairness, an emerging research thread advocates welfare-based fairness in ML to seek better compatibility between fair ML and distributive justice. This is in line with our proposal of using social welfare functions to capture a broader perspective on fairness. We next discuss a few representative papers in this thread, and review their chosen utility and social welfare definitions. \cite{liu2018delayed} look into the delayed impacts of fairness in machine learning on the welfare of the involved people. Using loan processing as the main running example, they suppose people have a performance variable, such as, credit score, that they would like to improve overtime. The paper's main results demonstrate that inserting fairness, which aims to benefit certain protected groups, in machine learning model does not guarantee long-term improvements for the targeted groups. In their analysis, they define utility functions based on the performance variables and individuals' expected outcomes, and use a utilitarian sum of individual utilities as the central decision maker's social welfare objective. \cite{heidari2018fairness} consider a standard supervised learning setting with true labels $\{y_i\}$ and predicted labels $\{\hat{y}_i\}$. They define the utility function as a function of $y_i, \hat{y}_i$, and the specific format is chosen to reflect whether $i$ is risk averse, neutral or seeking, and how close the predicted outcome $\hat{y}_i$ is to $i$'s desirable outcome. They then define a utilitarian sum of these individual utilities as the social welfare measure, and propose to add a constraint on this social welfare value to standard ML models as an in-processing fair ML approach. \cite{hu2020fair} study a similar utility definition without the risk component in a classification setup. They evaluate the overall welfare associated with classification decisions through comparing a vector of welfare values, which measure the utilitarian welfare by group. Also in a classification setting, \cite{corbett2018measure} suppose each group has fixed benefits and costs associated with classification outcomes, and these values are used as parameters in the utility functions. A group's utility aggregates the benefits and costs that individuals of the group incur from their classification outcomes. A more refined view of utility is studied in \cite{heidari2019moral}: they partition one's actual utility into an effort-based component and an advantage component. Utilizing this partition, they group individuals by effort-based utilities and propose a fairness measure equivalent to the expected advantage utility of the worst-off group.

\section{A Sampling of Social Welfare Functions} \label{sec:swfdef}
We briefly review a collection of SWFs to illustrate how they can embody various conceptions of equity.  For each, we indicate the type of optimization model it yields, and whether it is appropriate for our running example of mortgage loan processing.  We classify the SWFs as pure fairness metrics, functions that combine fairness and efficiency, and statistical fairness metrics.

\subsection{Pure fairness measures}

Social welfare functions that measure fairness alone, without an element of efficiency, are of two basic types: inequality metrics and fairness for the disadvantaged.  

Inequality metrics abound in the economics literature.  Some simple ones are represented by the following SWFs (which negate the inequality measure): 
\[
W(\vu)=
\left\{
\begin{array}{ll}
{\ds -(1/\bar{u})(u_{\max}-u_{\min})} & \mbox{for the {\em relative range}} \vsp \vsp \vsp \\
{\ds -(1/\bar{u})\sum_i |u_i-\bar{u}|} & \mbox{for the {\em relative mean deviation}} \vspace{-0.5ex} \\
{\ds -(1/\bar{u})\Big[(1/n)\sum_i(u_i-\bar{u})^2\Big]^{\half}} & \mbox{for the {\em coefficient of variation}} 
\end{array}
\right.
\]
There is also the well-known {\em Gini coefficient}, which is proportional to the area between the Lorenz curve and a diagonal line representing perfect equality.  It corresponds to the SWF 
\[
W(\bu)= 1-\frac{1}{2\bar{u}n^2} \sum_{i,j} |u_i-u_j|
\]

Other fairness-based SWFs are concerned with the lot of the disadvantaged.  The {\em Hoover index} measures the fraction of total utility that would have to be transferred from the richer half of the population to the poorer half to achieve perfect equality. The SWF is
\[
W(\bu) = -\frac{1}{2 n\bar{u}}\sum_i |u_i-\bar{u}|
\]
The Hoover index is proportional to the relative mean deviation and can therefore be optimized using the same model.

The {\em McLoone index} compares the total utility of individuals at or below the median utility to the utility they would enjoy if all were brought up to the median utility.  The index is 1 if nobody's utility is strictly below the median and approaches 0 if there is a long lower tail.  The SWF is 
\[
W(\bu) = \frac{1}{|I(\bu)|\tilde{u}} \sum_{i\in I(\bu)} \hspace{-1ex}  u_i
\]
where $\tilde{u}$ is the median of utilities in $\bu$ and $I(\bu)$ is the set of indices of utilities at or below the median.  

The Hoover and McLoone indices measure only the relative welfare of disadvantaged parties, and not their absolute welfare.  The {\em maximin} criterion addresses both.  It is based on the Difference Principle of John Rawls, which states that inequality should exist only to the extent it is necessary to improve the lot of the worst-off (\cite{Raw99,Fre03,RicWei99}).  It can be plausibly extended to a lexicographic maximum principle.  The SWF is simply
\[
W(\vu) = \min_i\{u_i\}
\]
Purely fairness-oriented SWFs can be used when equity is truly the only issue of concern.  In particular, they might be unsuitable for the mortgage problem, where overall utility is a prime consideration.

\subsection{Combining fairness and efficiency} \label{sec:combine}

Several SWFs combine equity and efficiency, sometimes with a parameter that regulates the relative importance of each.  Perhaps the best known is {\em alpha fairness}, for which the SWF is
\[
W_{\alpha}(\bu) = \left\{
\begin{array}{ll}
{\displaystyle
\frac{1}{1-\alpha}\sum_i u_i^{1-\alpha} 
} & \mbox{for}\; \alpha\geq 0, \; \alpha\neq 1 \vspace{0.5ex} \\
{\displaystyle
\sum_i \log(u_i) 
} & \mbox{for} \;\alpha=1
\end{array}
\right.
\]
Larger values of $\alpha$ imply a greater emphasis on equity, with $\alpha=0$ corresponding to a pure utilitarian criterion $\sum_i u_i$, and $\alpha=\infty$ to a pure maximin criterion.  An important special case is $\alpha=1$, which corresponds to {\em proportional fairness}, also known as the {\em Nash bargaining solution}.  It is widely used in telecommunications and other engineering applications.  Both proportional fairness and alpha fairness have been given axiomatic and bargaining justifications (\cite{Nas50,Har77,Rub82,BinRubWol86,LanKaoChiSab2010}).  

The {\em Kalai-Smorodinsky} (K--S) bargaining solution, proposed as an alternative to the Nash bargaining solution, minimizes each person's relative concession.  That is, it provides everyone the largest possible utility relative to the maximum one could obtain if other players are disregarded, subject to the condition that all persons receive the same fraction $\beta$ of their maximum.  In addition to the bargaining justification of \cite{KalSmo75}, this approach has been defended by \cite{Tho94} and is implied by the contractarian philosophy of \cite{Gau87}.  The SWF can be formulated 
\[
W(\bu) = \left\{
\begin{array}{ll}
\sum_i u_i, & \mbox{if $\bu=\beta\bu^{\max}$} \hspace{1ex} \mbox{for some $\beta$ with $0\leq\beta\leq 1$} \vspace{0.5ex} \\
0, & \mbox{otherwise}
\end{array}
\right.
\]
where $u_i^{\max} = \max_{(\vx,\vu) \in S_{\vec{x}\bu}} u_i$ for each $i$.

\cite{WilliamsCookson2000} suggest two {\em threshold} criteria for combining maximin and utilitarian objectives in a \mbox{2-person} context.  One uses maximin until the cost of fairness becomes too great, whereupon it switches to utilitarianism, and the other does the opposite.  \cite{HooWil12} generalize the former to $n$ persons by proposing the following SWF:
\[
W_{\Delta}(\vu) = (n-1)\Delta + \sum_{i=1}^n \max \big\{u_i-\Delta, u_{min}\big\} 
\]
where $u_{\min}=\min_i\{u_i\}$.  The parameter $\Delta$ regulates the equity/efficiency trade-off in a way that may be easier to interpret in practice than the $\alpha$ parameter: parties whose utility is within $\Delta$ of the lowest utility receive special priority.  Thus the disadvantaged are favored, and $\Delta$ defines who is disadvantaged.  As with the $\alpha$ parameter, $\Delta=0$ corresponds to a purely utilitarian criterion and $\Delta=\infty$ to a maximin criterion. 

One weakness of the model is that the actual utility levels of disadvantaged parties other than the very worst-off have no effect on the measurement of social welfare, as long as those utilities are within $\Delta$ of the lowest.  As a result, the socially optimal solution may not be as sensitive to equity as one might desire.  \cite{CheHoo20,CheHoo22a} address this issue by {\em combining utilitarianism with a leximax} rather than a maximin criterion.  A leximax (lexicographic maximum) solution is one is which the smallest utility is maximized, then subject to this value the second smallest is maximized, and so forth.  Chen and Hooker combine leximax and utilitarian criteria by maximizing a sequence of threshold SWFs, thereby obtaining more satisfactory solutions.

\subsection{Statistical bias metrics}

While we argue that bias metrics afford an overly narrow perspective on fairness, they nonetheless can be expressed as SWFs if desired.  The utility vector $\vu$ becomes simply a binary vector in which $u_i=1$ if individual $i$ is selected for some benefit, and $u_i=0$ otherwise.  We set constant $a_i=1$ when person $i$ actually qualifies for selection (as for example when person $i$ in the mortgage training set repaid the loan), and $a_i=0$ otherwise.  Two groups are compared, respectively indexed by $N$ and $N'$.  One is a protected group, such as a minority subpopulation, and the other consists of the rest of the population.  

For example, {\em demographic parity} has the SWF
\[
W(\vu) = 1 - \left|
\frac{1}{|N|}\sum_{i\in N} u_i - \frac{1}{|N'|} \sum_{i\in N'} u_i
\right|
\]
{\em Equalized odds} can be measured in two ways, one of which is {\em equality of opportunity}:
\[
W(\vu) = 1 - \left|
\frac{\sum_{i\in N} a_iu_i}{\sum_{i\in N} a_i}
- \frac{\sum_{i\in N'} a_iu_i}{\sum_{i\in N'} a_i}
\right|
\]
and still another {\em predictive rate parity}:
\[
W(\vu) = 1 - \left|
\frac{\sum_{i\in N} a_iu_i}{\sum_{i\in N} u_i} -
\frac{\sum_{i\in N'} a_iu_i}{\sum_{i\in N'} u_i} 
\right|
\]

Bias measures take no account of efficiency.  One can, of course, maximize predictive accuracy subject to constraints on the amount of bias, but this has a number of drawbacks:

\begin{itemize}
    \item As previously argued, it provides a very limited perspective on the utility actually created by decisions.  Indeed, the utility vector consists only of 0--1 choices.
    \item Bias measurement forces one to designate ``protected groups'' (as indicated by the index set $N$).  There is no clear principle for selecting which groups should be protected, unless one is content simply to recognize those mandated by law. 
    \item There is no consensus on which bias measure is suitable in a given context, if any.  Bias measures were developed by statisticians to measure predictive accuracy, not to assess fairness.
    \item There is no principle for balancing equity and efficiency.  If equity is part of the objective function, as in social welfare maximization, the choice of that function obliges one to justify the equity/efficiency trade-off mechanism in a transparent manner.  
    \item Bias measurement forces one to identify {\em a priori} which individuals in a training set should be selected for benefits (as indicated by $a_i$).  In a social welfare approach, no prior decisions of this kind are necessary.   
\end{itemize}

\section{Case Study: Mortgage Loan Processing}
We demonstrate the proposed framework on our running example. Specifically, we develop loan decision models integrated with machine learning for social welfare optimization. Recall from Section \ref{sec:ex-loan}, a bank faces the task to process loan applications and allocate available funds to the approved applicants. For this task, the stakeholders are loan applicants. The bank has a conventional efficiency-driven goal to approve qualified applicants and avoid loan defaults. Additionally, the bank aims to achieve a fair distribution of utilities among the loan applicants. 

\subsection{Decision models}
Suppose an applicant has a feature profile $\vx_i$ and a true label $y_i \in \{1,-1\}$ indicating his/her qualification status, namely, whether he/she will default on an approved loan. The bank has data on past loan processing decisions and repay outcomes from applicants of mixed profiles. To inform loan decisions, the bank can train ML models based on historical loan processing data to predict applicants' default risks. 

We consider logistic regression, one of the popular ML methods in practice. A standard logistic regression model is trained through the following loss minimization problem, where $\vtheta$ is the decision boundary.
\begin{gather} \label{sys:lr-uc}
    \begin{aligned}
    \min_{\vtheta} ~ C\sum_i \log(1+e^{-y_i\langle \vtheta, \vx_i\rangle}) + \lambda_1 \norm{\vtheta}_1
    \end{aligned}
\end{gather}
\noindent
Based on this model, $\frac{1}{1+e^{-\langle \vtheta, \vx_i \rangle}}$ can be interpreted as the probability of $y_i = 1$, which we use as the predicted probability for an applicant with the features $\vx_i$ to repay a loan. To construct a classifier from the logistic regression model, we use a threshold $\tau \in [0,1]$ on these output probabilities. When $\frac{1}{1+e^{-\langle \vtheta, \vx_i \rangle}} > \tau$, $i$ will be labeled positive, i.e. $\hat{y}_i = 1$; otherwise, the predicted label is negative, $\hat{y}_i = -1$. The typical threshold is $\tau = 0.5$, and increasing the threshold corresponds to a more strict positive classification standard. 

\subsubsection*{Post-processing integration}
In this integration scheme, we utilize the predicted repay probabilities $\hat{p}_i = \frac{1}{1+e^{-\langle\vtheta,\vx_i\rangle}}$ in loan processing. Based on $\{\hat{p}_i\}$, we use the following model to optimize the social welfare in loan decisions.
\begin{gather} \label{sys:dec}
    \begin{aligned}
    \min_{\vu, \vd} ~ & W(\vu) \\
   \text{s.t. } & u_i = \hat{p}_i d_i, d_i \in [0, r_i] ~\forall i; ~ \sum_i d_i \leq B.
    \end{aligned}
\end{gather}
As discussed in Section \ref{sec:step3}, an applicant's expected utility $u_i$ from the loan decision is dependent on the prediction $\hat{p}_i$. The utility definition characterizes that $i$'s utility from receiving a loan of amount $d_i$ is exactly $d_i$ if $i$ could repay the loan, but is $0$ if $i$ would default.

\subsubsection*{In-processing integration}
In this approach, social welfare optimization is integrated directly into the prediction/decision model. The additional hyperparameter $\lambda_2 \geq 0$ regulates the importance level of $W(\vu)$ in the training phase. 
\begin{gather} \label{sys:lr-inte}
    \begin{aligned}
    \min_{\vtheta, \vu} ~ & C\sum_i \log(1+e^{-y_i\langle \vtheta, \vx_i\rangle}) + \lambda_1 \norm{\vtheta}_1 - \lambda_2 W(\vu) \\
   \text{s.t. } & u_i = g_{y_i} \hat{y}_i + b_{y_i}, ~\forall i \\
    & \hat{y}_i = 1 \text{ if } \langle \vtheta, \vx_i \rangle \geq 0, \hat{y}_i = 0 \text{ otherwise}, ~\forall i.
    \end{aligned}
\end{gather}
In this training problem, each involved applicant receives a predicted label $\hat{y}$ as the decision. We assume a linear utility function $u_i(\hat{y}_i) = g_i \hat{y}_i + b_i$, where $g_i, b_i$ are assigned based on the true label $y_i$. Specifically, we assign $g_1=0.25, g_{-1}=-0.25$ and $b_1 = 0.5, b_{-1}=0.25$.
An applicant has four possible outcomes from the prediction task: true positive (TP) when a qualified applicant is correctly classified, and similarly false positive (FP), true negative (TN), false negative (FN). With the assigned parameters, the four outcomes have the following utilities: $u^{TP} = 0.75, u^{TN}=0.5, u^{FP}=0, u^{FN}=0.25$. As shown, we have defined utilities to let correct predictions provide greater benefits than incorrect predictions, which fits the bank's efficiency-driven goal to identify qualified applicants reliably. Additionally, we set a FP outcome to be more costly for a disqualified applicant than a FN outcome for a qualified applicant. 

For both integrated models, we consider the following social welfare definitions: utilitarian SWF $W(\vu) = \sum_i u_i$, alpha fairness $W(\vu) = 2\sum_i u_i^{1/2}$ at $\alpha = 0.5$, proportional fairness $W(\vu) = \sum_i \log{u_i}$ and the maximin criterion $W(\vu) = \min_i u_i$. We have discussed in Section \ref{sec:swfdef} that these SWFs reflect different positions on the spectrum from pure efficiency to pure fairness. Our experiment results will illustrate that the SWF choices impact decisions' welfare. 


\subsection{Dataset and implementation details}
We work with a small scale instance based on the German credit dataset (\cite{german}). This dataset is a common choice for investigating fair machine learning models. It contains $1000$ entries of people labeled as having good ($y_i = 1$) or bad ($y_i = -1$) credit risks. Each entry has $20$ attributes describing relevant information including age, gender, employment status, etc. For the purpose of our numerical study, we view these people as loan applicants. Those with good credit risks are qualified applicants who will repay a loan, and the others are unqualified and will default. 

On this dataset, we use a 80/20 training-testing split, namely, 800 entries are used to train a logistic regression model, and the remaining 200 entries form the testing dataset and represent the new loan applicants to process. For each person $i$ in the testing dataset, we randomly generate the requested loan amount $d_i \in [0,100]$. The bank's available fund is $B=5000$. We repeat all experiment instances on $5$ randomly generated training-testing splits and refer to the average statistics for analysis. In the training models, we fix $C=1, \lambda_1=10^{-4}$ and test a range of values for $\lambda_2 \in \{0.1,1,10,50,100\}$ for the in-processing integration. While more advanced hyperparameter tuning on all three parameters could be useful for improving the predictive accuracy, we adopt the simpler parameter setups to focus on understanding the impacts of the social welfare optimization.

We solve the standard logistic regression model using the scipy minimize function with the 'SLSQP' method. We solve all the other optimization models 
using Gurobi 9.1.2. All codes are written in Python 3.8 on a desktop PC running Windows 10 Pro 64-bit with Intel(R) Core(TM) i7-7600U CPU @ 2.80GHz processors and 24 GB of RAM. 

\subsection{Results and findings}
From examining the dataset, we notice that the labels are unbalanced across age ranges. For example, among the $548$ individuals younger than $35$ years old, 65\% have a positive true label (i.e. they are viewed as qualified applicants), whereas 76\% of the 452 older applicants are labeled positive. 
In the AI Fairness 360 Toolkit, \cite{bellamy2018ai} identify the older individuals as more privileged in ML models trained using the German credit dataset. We first verify this observation in the standard logistic regression models. 
\begin{table}
    \centering
    \begin{tabular}{c|c|c|c} \toprule
    group & positive rate (\%) & true positive rate (\%) & true negative rate (\%) \\ \hline
      young &  70.33 $\pm$ 2.76 & 82.03 $\pm$ 3.74 & 51.05 $\pm$ 5.23 \\ \hline
      old   &  82.14 $\pm$ 3.21 & 91.01 $\pm$ 4.61 & 45.27 $\pm$ 3.39 \\ \bottomrule
    \end{tabular}
    \caption{\small{Classification outcomes for young and old groups on test dataset. Reported numbers are means and standard deviations from $5$ random train-test splits. 
    }}
    \label{tab:lr-outcome}
\end{table}

As shown in Table \ref{tab:lr-outcome}, the standard prediction model without fairness considerations violates demographic parity and equality of opportunity against the young group. We next demonstrate that applying social welfare optimization with both integration methods could improve group parity through seeking welfare-based fairness. 

\begin{figure}[h]
    \centering
    \begin{subfigure}{0.8\textwidth}
    \centering
    \includegraphics[width=\textwidth]{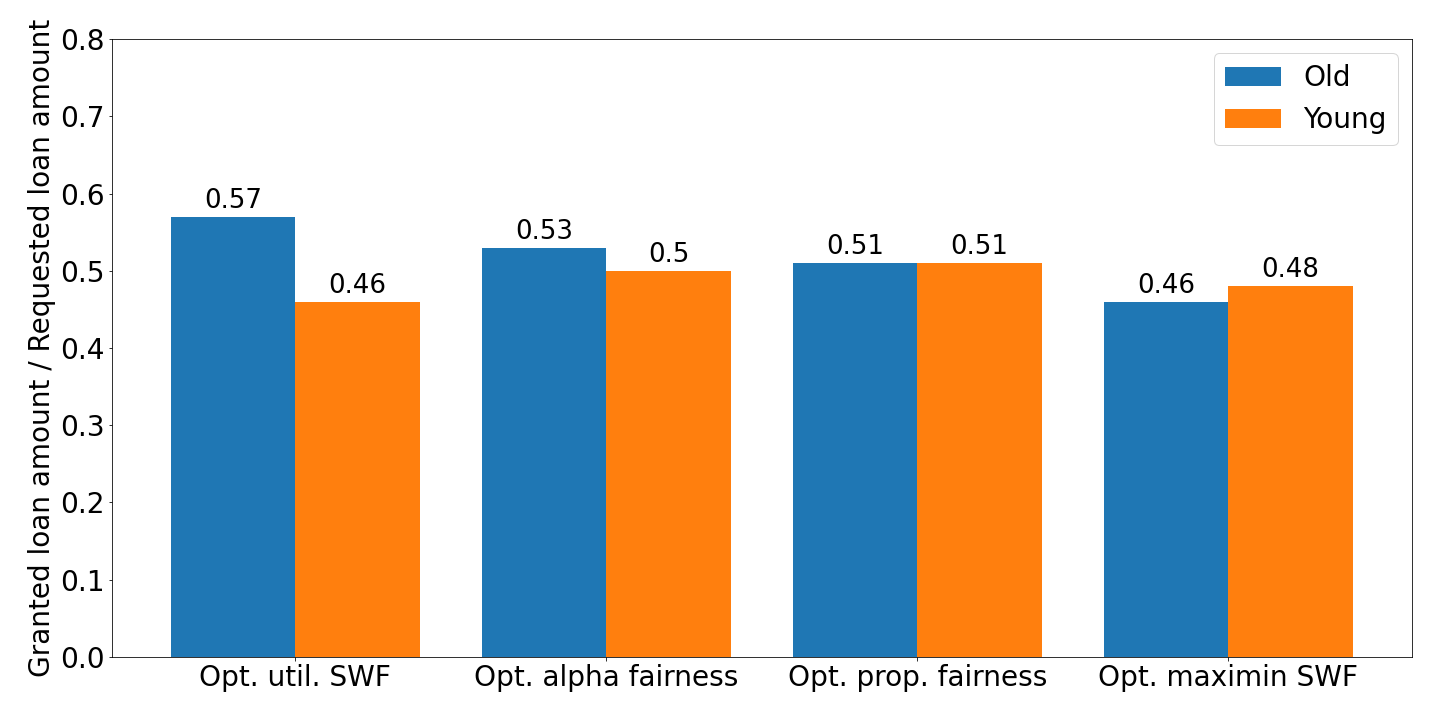}
    \vspace{-5ex}
    \caption{Approval ratios for all applicants}
    \label{fig:app-ratio}
    \end{subfigure}
        \begin{subfigure}{0.8\textwidth}
        \centering
    \includegraphics[width=\textwidth]{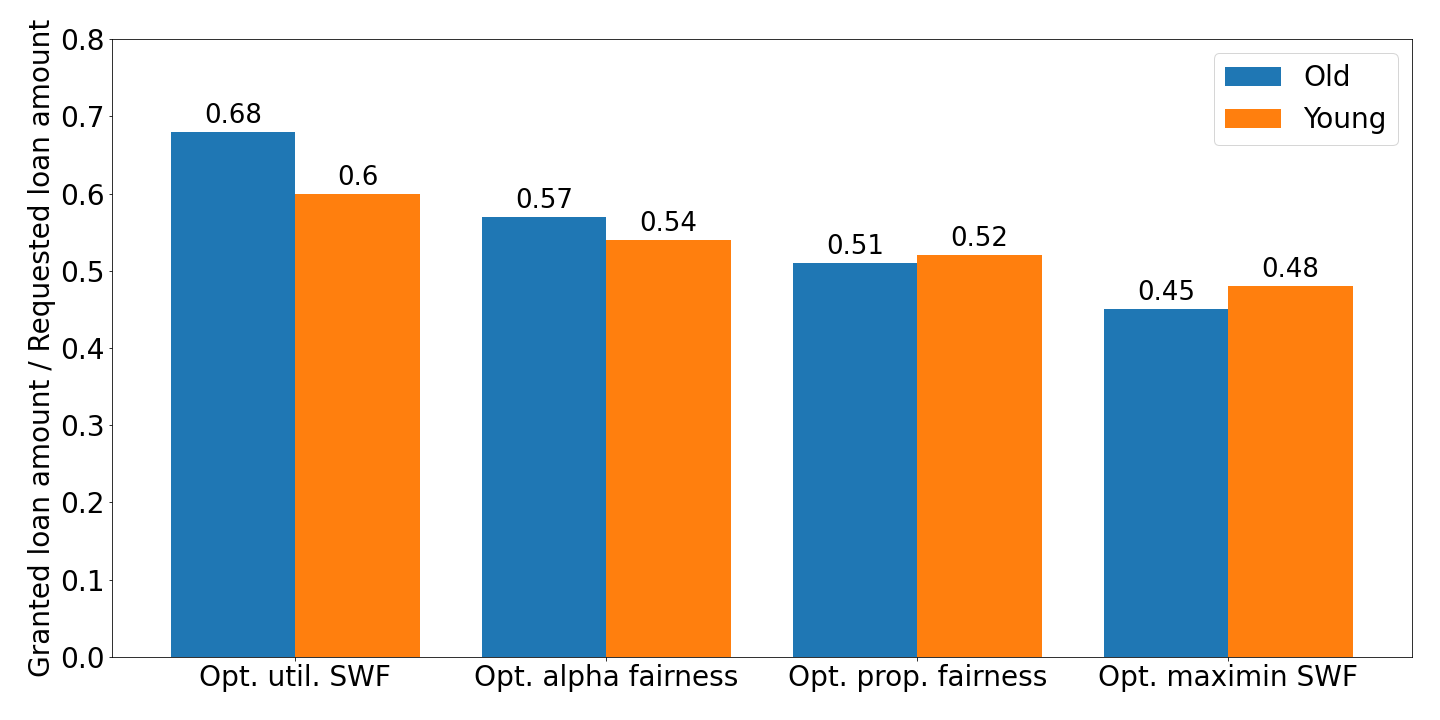}
        \vspace{-5ex}
    \caption{Approval ratios for qualified applicants}
    \label{fig:app-qual-ratio}
    \end{subfigure}
    \caption{\small{Loan distributions for young and old groups from post-integration with four SWF objectives. Reported numbers are means from 5 random test datasets.}}
    \label{fig:util-dist}
\end{figure}

In the post-processing method, we focus on comparing the loan distributions to young and old applicants. We consider two performance metrics: the ratio between the total received loans and the total requested loans by applicants in either group, and the ratio computed based on only the qualified applicants. Since the old group is more privileged in the prediction task, the applicants in this group are better positioned for loan distribution if the objective is primarily efficiency driven. For example, Fig. \ref{fig:util-dist} shows that optimizing a utilitarian SWF leads to higher approved ratios for the old group. On the contrary, through optimizing a SWF with greater emphasis on fairness, e.g. proportional fairness, the obtained loan distributions are fairer between groups. We also note that improving fairness costs efficiency, as the overall loan approval ratios are lower than the utilitarian solutions.

\begin{figure}
    \begin{subfigure}{0.8\textwidth}
    \centering
    \includegraphics[width=\textwidth]{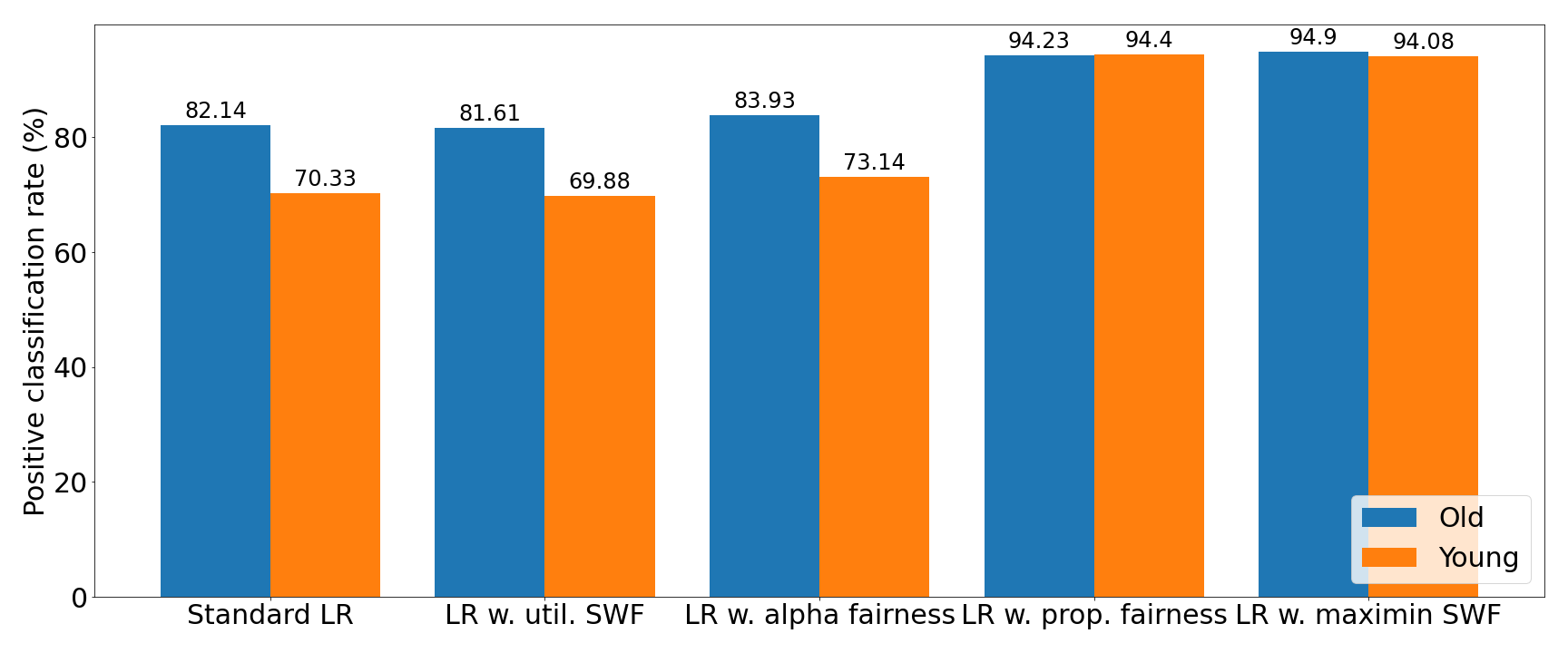}
        \vspace{-5ex}
    \caption{Positive rates (\%), $\lambda_2 = 0.1$}
    \label{fig:pos0}
    \end{subfigure}
    \centering
    \begin{subfigure}{0.8\textwidth}
    \centering
    \includegraphics[width=\textwidth]{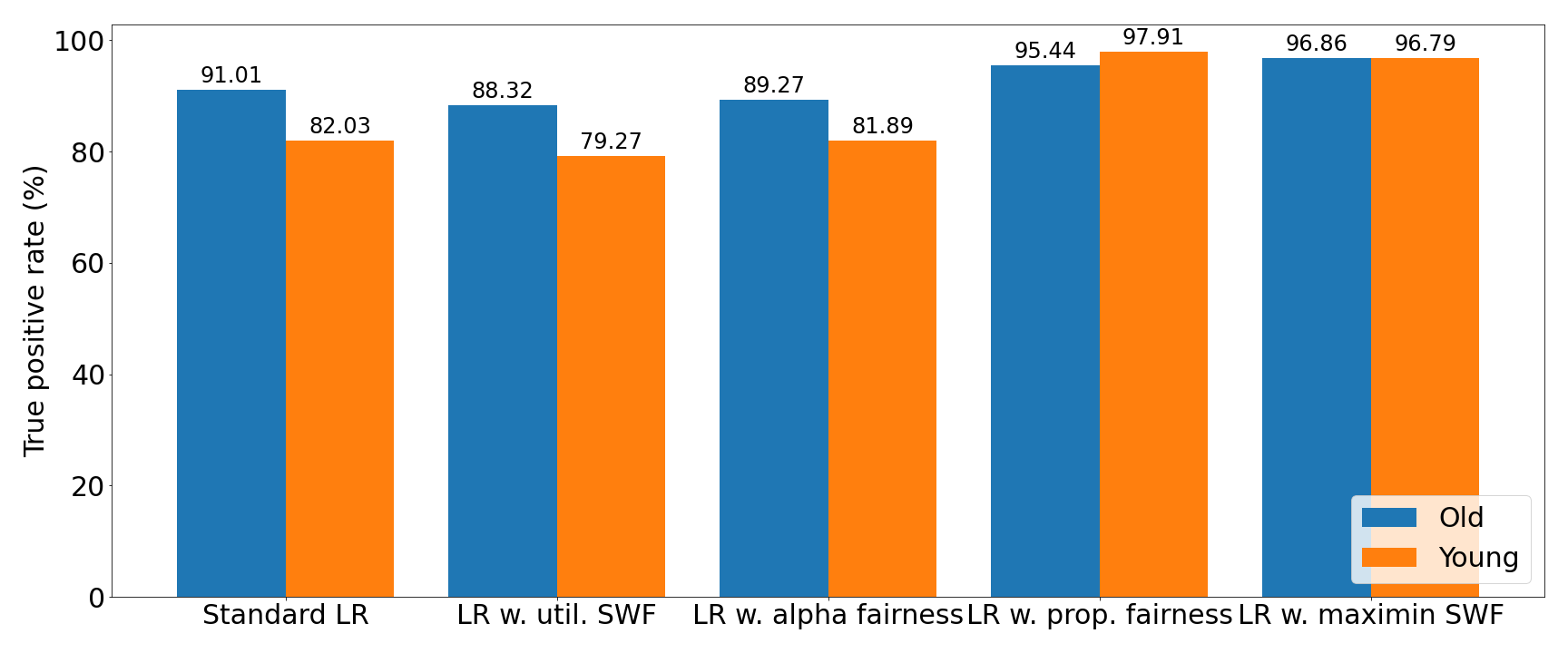}
        \vspace{-5ex}
    \caption{True positive rates (\%), $\lambda_2 = 0.1$}
    \label{fig:tp0}
    \end{subfigure}
    \begin{subfigure}{0.8\textwidth}
    \centering
    \includegraphics[width=\textwidth]{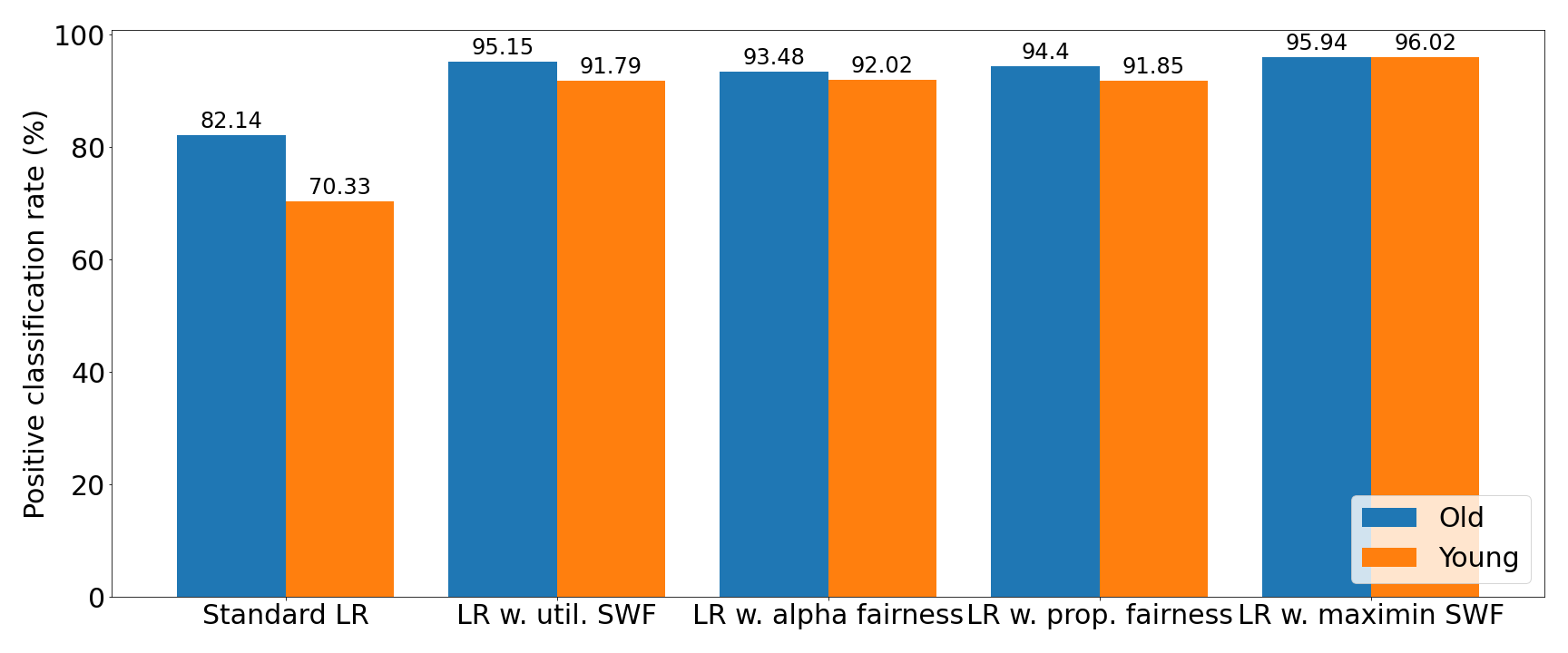}
        \vspace{-5ex}
    \caption{Positive rates (\%), $\lambda_2 = 1$}
    \label{fig:pos1}
    \end{subfigure}
    \centering
    \begin{subfigure}{0.8\textwidth}
    \centering
    \includegraphics[width=\textwidth]{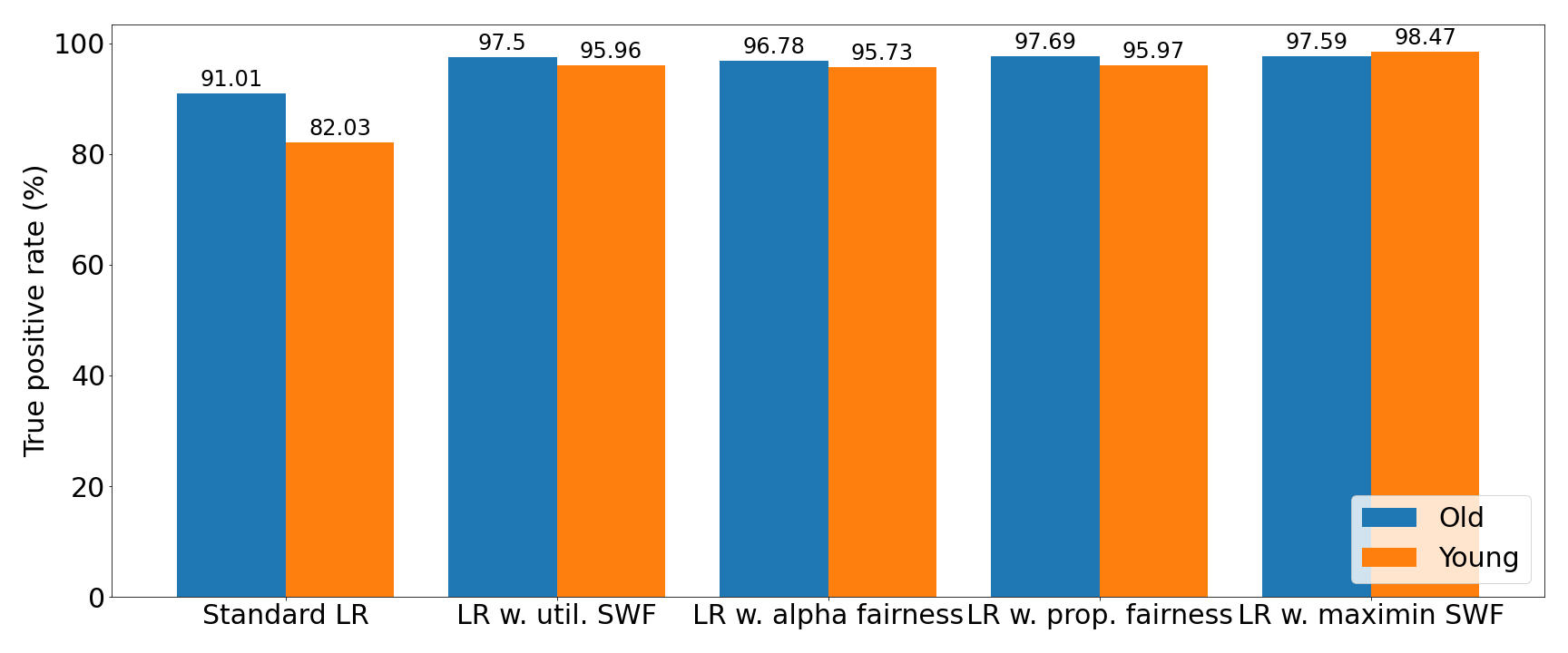}
        \vspace{-5ex}
    \caption{True positive rates (\%), $\lambda_2 = 1$}
    \label{fig:tp1}
    \end{subfigure}
    \caption{\small{Classification outcomes for young and old groups on test dataset using standard logistic regression and in-processing integrated models with four SWFs. Reported numbers are means from 5 random train-test splits. The performances at $\lambda_2 \in \{10,50,100\}$ are highly similar to those at $\lambda_2=1$ shown in (c), (d).}}
    \label{fig:pred-perf}
\end{figure}
\begin{figure}[h]
    \centering
    \includegraphics[width=1\textwidth]{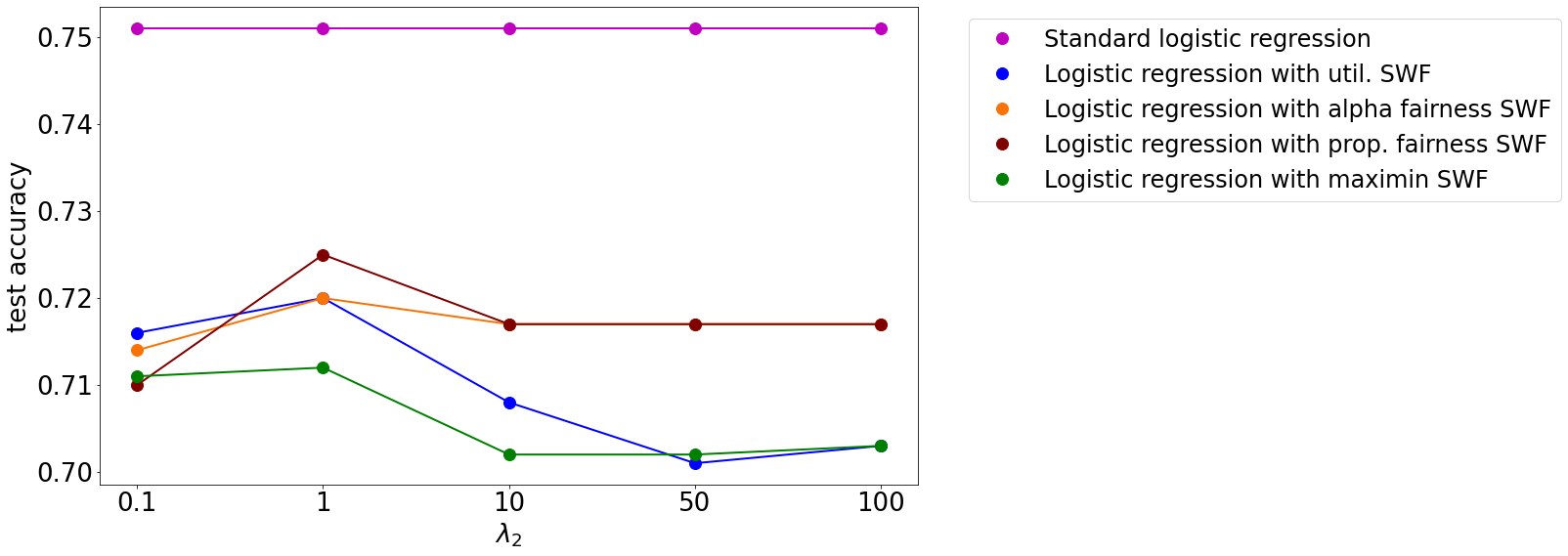}
    \caption{\small{Test accuracy over different strengths of social welfare integration. Plots are generated based on the means from $5$ random train-test splits. The choices of $\lambda_2$ have small impacts on test accuracy. Among the social welfare definitions, integrated models with alpha fairness and proportional fairness have slightly higher test accuracy.}}
    \label{fig:acc} 
\end{figure}

In the in-processing method, we compare the positive classification rates and true positive rates from classification. Figs. \ref{fig:pos0} and \ref{fig:tp0} illustrate that, when social welfare optimization has a small weight in training, optimizing proportional fairness and the maximin criterion leads to better fairness outcomes for both groups, whereas the other two options only slightly reduce the disparities. As the social welfare weight increases, all the SWF choices provide effective improvement in group parities. Additionally, we note that including social welfare optimization in the training problem reduces the test accuracy by a small amount (see Fig. \ref{fig:acc}). 

\section{Discussion and Conclusion} \label{sec:conclusion}
We formalize a general framework for using optimization to incorporate welfare-based fairness into AI applications. The framework provides a guideline for formulating a decision task into a social welfare optimization problem. In particular, we illustrate how optimization can be integrated with rule-based AI systems and ML models. By expanding the fairness problem to the optimization of social welfare functions, one can achieve a broader perspective on fairness that are driven by the well-beings of stakeholders and characterize the broader fairness concepts in a principled way. 
On a small scale loan processing application, we demonstrate that decision models integrating social welfare optimization with ML could improve fairness among groups without requiring pre-specified group labels as protected vs. unprotected.  

We conclude the paper by outlining a brief research program to explore some key questions related to the framework. 
\begin{itemize}
    \item The presented general framework opens up numerous questions. For integration with rule-based AI, one important direction is to investigate how to build ethics-sensitive rule bases to fit into different social welfare optimization scenarios. Previous works on formulating ethics principles into rules, e.g. \cite{BriArkBel06,KimHooDon2021}, may provide guidance for this direction. For integration with ML, future research could 
    explore the empirical direction further on large scale and real-world applications, which could also provide engagement opportunities with practitioners to specify context-specific utility and social welfare definitions. Theoretical connections between social welfare optimization and group parity improvement could be useful to explore as well.
    
    \item Although optimization solvers have been developed over decades, not all classes of optimization models are readily solvable by state-of-the-art software. 
    For practical use of social welfare optimization models, one may need to apply available computational strategies or design problem-specific heuristics to speed up solving the optimization problems. 
    
    \item The social welfare functions we consider are of a static nature, that is, a SWF does not attempt to capture potential dynamics in the utilities. 
    A dynamic perspective may be required in sequential decision problems where decisions need to be made repeatedly and the selected actions have incremental impacts on the long term social welfare. Future research could explore how to extend the presented optimization based framework to fit a dynamic view of welfare and fairness. Although this is not a trivial task, there are many well-developed techniques to utilize, such as, stochastic optimization, Markov decision process, etc. 
\end{itemize}

\bibliographystyle{abbrvnat}
\bibliography{ref}
\end{document}